\pdfoutput=1
\documentclass[conference]{IEEEtran}
\IEEEoverridecommandlockouts

\usepackage{cite}
\usepackage{amsmath,amssymb,amsfonts}
\usepackage{algorithmic}
\usepackage{graphicx}
\usepackage{textcomp}
\usepackage{xcolor}
\usepackage{booktabs}
\usepackage{multirow}
\usepackage{placeins}

\makeatletter
\def\@citex[#1]#2{%
  \leavevmode
  \@beginparpenalty\@lowpenalty
  \let\@citea\@empty
  [\@for\@citeb:=#2\do
    {\@citea\def\@citea{, }%
     \edef\@citeb{\expandafter\@firstofone\@citeb\@empty}%
     \if@filesw\immediate\write\@auxout{\string\citation{\@citeb}}\fi
     \@ifundefined{b@\@citeb}{\mbox{\reset@font\bfseries ?}%
       \G@refundefinedtrue\@latex@warning
       {Citation `\@citeb' on page \thepage\space undefined}}%
       {\hbox{\csname b@\@citeb\endcsname}}%
    }]
  \penalty\@highpenalty}
\makeatother

\def\BibTeX{{\rm B\kern-.05em{\sc i\kern-.025em b}\kern-.08em
    T\kern-.1667em\lower.7ex\hbox{E}\kern-.125emX}}

\begin{document}

\title{ScatSPADE: Scattering-Aware Spatially-Adaptive Normalization for Optical-to-SAR Image Translation}

\author{
	\begin{minipage}[t]{0.32\textwidth}
		\centering
		1\textsuperscript{st} Zhenghao Zhang\\
		\textit{College of Electronic Science and Technology}\\
		\textit{National University of Defense Technology}\\
		Changsha, China\\
		zhangzhenghao25@nudt.edu.cn
	\end{minipage}%
	\hfill
	\begin{minipage}[t]{0.32\textwidth}
		\centering
		2\textsuperscript{nd} Xinjie Wang\\
		\textit{College of Electronic Science and Technology}\\
		\textit{National University of Defense Technology}\\
		Changsha, China\\
		wangxinjie@nudt.edu.cn
	\end{minipage}%
	\hfill
	\begin{minipage}[t]{0.32\textwidth}
		\centering
		3\textsuperscript{rd} Binwei Hui\\
		\textit{College of Electronic Science and Technology}\\
		\textit{National University of Defense Technology}\\
		Changsha, China\\
		huibingwei07@nudt.edu.cn
	\end{minipage}%
	\\[1em]
	\hspace{0.15\textwidth}
	\begin{minipage}[t]{0.32\textwidth}
		\centering
		4\textsuperscript{th} Jiyuan Liu\\
		\textit{College of Electronic Science and Technology}\\
		\textit{National University of Defense Technology}\\
		Changsha, China\\
		liujiyuan@nudt.edu.cn
	\end{minipage}%
	\hfill
	\begin{minipage}[t]{0.32\textwidth}
		\centering
		5\textsuperscript{th} Wei Wang\textsuperscript{*}\\
		\textit{College of Electronic Science and Technology}\\
		\textit{National University of Defense Technology}\\
		Changsha, China\\
		wwang@nudt.edu.cn
	\end{minipage}%
	\hspace{0.15\textwidth}
}

\maketitle

\begin{abstract}
We address unpaired optical-to-SAR image translation, which augments the scarce labeled SAR data used for target recognition and scene understanding by synthesizing SAR-like images from abundant optical data. Existing methods typically treat this translation as a generic style-transfer problem and ignore the electromagnetic scattering mechanisms that distinguish SAR from optical imagery, leaving region-specific scattering structure unmodeled. We propose ScatSPADE, a framework that injects a self-supervised scattering prior into the generator through two complementary, scattering-conditioned modulation modules. The key to our approach is a per-pixel scattering map predicted from generator bottleneck features, which drives both a Scattering-Aware Spatially-Adaptive Normalization (SPADE) that replaces uniform decoder normalization and a Scattering-Aware Attention (SAA) module that re-weights bottleneck features toward scattering-critical regions. On the QXS-SAROPT dataset, ScatSPADE reduces FID to 22.62, a 57\% reduction over a U-GAT-IT base architecture and below the paired pix2pix upper bound (62.92), while producing properly grayscale SAR with LPIPS among the lowest of the unpaired methods. Notably, replacing the scattering map with a generic spatial prior raises FID, showing that scattering-aware conditioning yields a specific gain beyond spatially-adaptive normalization alone.
\end{abstract}

\begin{IEEEkeywords}
SAR image generation, image-to-image translation, scattering mechanism, spatially-adaptive normalization
\end{IEEEkeywords}


\section{Introduction}

Synthetic Aperture Radar (SAR) is an indispensable Earth-observation modality, providing all-weather, day-and-night imaging for reconnaissance, disaster monitoring, and terrain mapping \cite{ref1,ref5}. SAR imagery is, however, inherently difficult to interpret due to coherent speckle noise and non-intuitive scattering geometry \cite{ref3}, and the scarcity of labeled SAR data bottlenecks the deep-learning models behind modern target recognition and scene understanding \cite{ref7,ref9}. To bridge this annotation gap, optical-to-SAR image translation synthesizes SAR-like images from abundantly available optical data that are far easier to annotate, and has become a promising data-augmentation technique \cite{ref10}.

CycleGAN-based approaches \cite{ref16,ref23} dominate unpaired optical-to-SAR translation, and subsequent work improves fidelity through frequency-aware losses \cite{ref53}, cross-domain or residual attention \cite{ref24,ref25}, cascaded-residual generators \cite{ref34}, and wavelet or diffusion generators \cite{ref42,ref43}. These methods achieve impressive visual quality, yet they all cast translation as a statistical style-transfer problem and do not model the electromagnetic scattering physics that governs SAR image formation \cite{ref52}. Consequently, synthesized SAR often exhibits color casts, overly smooth speckle, and inaccurate point-like returns at man-made structures.

In reality, SAR imagery is governed by two dominant scattering modes---distributed scatterers (vegetation, soil, water) that produce granular speckle and strong scatterers (buildings, metallic objects) that produce intense point-like returns \cite{ref48,ref49,ref57,ref58}---so a faithful generator must modulate features very differently across these regions. We observe that existing generators fall short for two reasons: their uniform instance normalization applies a single affine transform across the entire feature map and cannot produce the region-specific intensity scaling that this bimodal radiometry demands; and they offer no scattering-aware conditioning signal, so neither their normalization nor their attention can adapt to where strong versus distributed scatterers occur in a scene.

To address both issues, we propose \textbf{ScatSPADE}, a framework that injects a self-supervised scattering prior into the generator. A lightweight bottleneck module first predicts a per-pixel, two-class scattering map from the generator's own features, without any manual annotation. The core of our approach is that this single map drives two complementary, scattering-conditioned modulation modules: (i)~Scattering-Aware SPADE, which replaces uniform normalization with spatially adaptive affine parameters so that strong- and distributed-scatterer regions receive different modulation, and (ii)~a Scattering-Aware Attention (SAA) module, which re-weights bottleneck features toward scattering-critical regions. Together they form a ``predict $\to$ attend $\to$ normalize'' pipeline that requires no manual annotation. Our contributions are as follows:
\begin{itemize}
\item We propose Scattering-Aware SPADE, which conditions decoder normalization on a self-supervised scattering map, enabling region-specific feature statistics without manual annotation.
\item We introduce a Scattering-Aware Attention (SAA) module guided by the same prior, and show that it is independently effective and complementary to SPADE.
\item On the QXS-SAROPT benchmark, ScatSPADE reduces FID to 22.62---a 57\% reduction over a U-GAT-IT base architecture and below the paired pix2pix upper bound---with properly grayscale, perceptually faithful SAR.
\end{itemize}


\section{Related Work}

\textbf{Unpaired image-to-image translation.} CycleGAN \cite{ref16} introduced cycle-consistency for unpaired bidirectional mapping; CUT \cite{ref47} simplified this to one-sided translation via contrastive learning over patches; U-GAT-IT \cite{ref31} used attention-guided adaptive layer-instance normalization; and SPADE \cite{spade} conditioned spatially-adaptive normalization on semantic segmentation maps. Among these, SPADE conditions spatial modulation on semantic segmentation maps for layout-to-image synthesis, yet none of them account for the electromagnetic scattering mechanisms that distinguish SAR from optical imagery.

\textbf{Optical-to-SAR image generation.} Cross-modal optical-SAR translation has attracted attention for SAR data augmentation \cite{ref10}. CycleGAN-based methods \cite{ref16,ref23} dominate; subsequent work improves fidelity via DCT-based spectral losses \cite{ref53}, cross-domain or residual attention \cite{ref24,ref25}, cascaded-residual generators \cite{ref34}, or wavelet \cite{ref42} and diffusion \cite{ref43} generators. All, however, treat translation as style transfer without the physical scattering taxonomy governing SAR.

\textbf{SAR scattering mechanisms.} SAR images are governed by electromagnetic scattering rather than optical reflectance \cite{ref52}: rough surfaces (vegetation, soil, water) act as distributed scatterers producing speckle \cite{ref3,ref48,ref49}, while man-made structures produce intense point-like returns \cite{ref57,ref58}. This distributed/strong dichotomy is central to SAR interpretation, yet existing optical-to-SAR methods do not model it in their feature modulation.

\FloatBarrier
\section{Proposed Method}

\begin{figure*}[!tbp]
\centerline{\includegraphics[width=\textwidth]{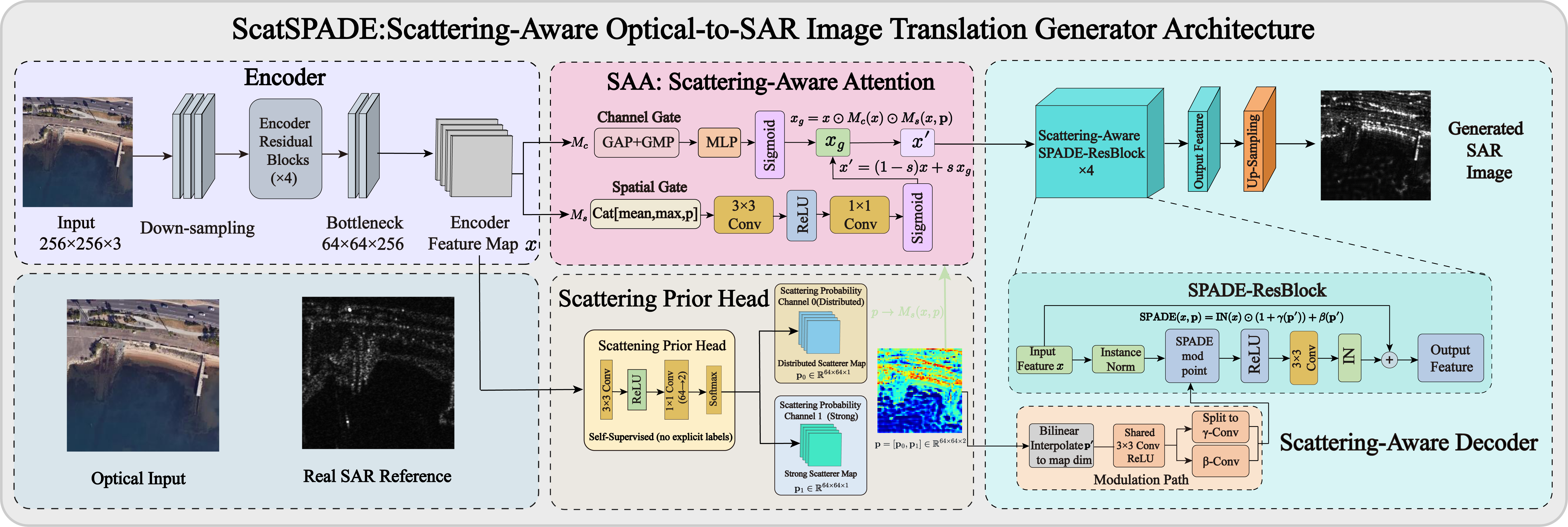}}
\caption{Overall architecture of ScatSPADE. The generator encodes the input optical image, predicts a per-pixel scattering map at the bottleneck, re-weights features through SAA, and decodes with scattering-conditioned SPADE normalization.}
\label{fig:framework}
\end{figure*}

\subsection{Overview}

As illustrated in Fig.~\ref{fig:framework}, we adopt the U-GAT-IT bidirectional architecture \cite{ref31}: two generators, $G_{A \to B}$ (optical$\to$SAR) and $G_{B \to A}$ (SAR$\to$optical), are trained jointly with cycle-consistency constraints for unpaired translation. This paper focuses on the forward generator $G_{A \to B}$, whose encoder downsamples the input to a $64 \times 64 \times 256$ bottleneck. A lightweight bottleneck module predicts a per-pixel, 2-class scattering map $\mathbf{p}$ from these bottleneck features. This single learned prior drives two complementary modulation modules: (i) a Scattering-Aware Attention (SAA) block that re-weights bottleneck features, and (ii) Scattering-Aware SPADE layers that replace all instance normalization in the decoder.

\subsection{Scattering Prior}
\label{sec:classifier}

SAR imagery is governed by the two dominant scattering modes introduced above. We model this dichotomy with a lightweight 2-class head at the generator bottleneck. Two convolutional layers map the $64 \times 64 \times 256$ feature to a map $\mathbf{p} \in \mathbb{R}^{64 \times 64 \times 2}$ via softmax, where $\mathbf{p}_0$ and $\mathbf{p}_1$ target the distributed and strong scatterer classes, respectively. The head receives no explicit scattering labels; it is trained implicitly through the downstream SPADE and SAA gradients. We treat this learned map as a self-supervised spatial conditioner rather than a calibrated scattering classifier: it does not faithfully predict hand-crafted scattering labels, yet conditioning SPADE and SAA on it yields substantially better realism than a generic, non-scattering spatial prior (Table~\ref{tab:ablation}). The map $\mathbf{p}$ serves as the shared conditioning signal for both modules below.

\subsection{Scattering-Aware Modulation}
\label{sec:spade}

Standard instance normalization (IN) applies a uniform affine transform across the entire feature map, ignoring the fact that strong-scatterer regions require different modulation than distributed-scatterer regions. We address this with two scattering-conditioned modules.

\textbf{Scattering-Aware SPADE.} We replace IN in the decoder with spatially adaptive normalization conditioned on $\mathbf{p}$:
\begin{equation}
    \text{SPADE}(\mathbf{x}, \mathbf{p}) = \text{IN}(\mathbf{x}) \odot (1 + \gamma(\mathbf{p}')) + \beta(\mathbf{p}')
    \label{eq:spade}
\end{equation}
where $\mathbf{p}'$ is the bilinearly upsampled scattering map, and $\gamma$, $\beta$ are generated by a shared convolutional MLP. This enables region-specific variance scaling and mean shifting: strong-scatterer regions can preserve sharp peaks, while distributed regions receive smoother modulation.

\textbf{Scattering-Aware Attention (SAA).} SPADE modulates feature statistics. We complement it with an attention module that modulates feature \emph{saliency}. Let $M_c$ and $M_s$ denote the channel and spatial gating maps; the fully gated feature is $\mathbf{x}_g = \mathbf{x} \odot M_c(\mathbf{x}) \odot M_s(\mathbf{x}, \mathbf{p})$. Rather than replacing $\mathbf{x}$ with $\mathbf{x}_g$ directly, we aggregate them through a learnable residual coefficient $s$ (initialized to $0.1$):
\begin{equation}
    \mathbf{x}' = \mathbf{x} + s\cdot(\mathbf{x}_g - \mathbf{x}) = (1-s)\,\mathbf{x} + s\,\mathbf{x}_g
    \label{eq:saa}
\end{equation}
This identity-preserving form is deliberate: because SAA sits at the bottleneck and feeds the SPADE decoder, a hard multiplicative gate ($s{=}1$) over-suppresses features and degrades the decoder's distributional realism, whereas the residual lets SAA \emph{complement} rather than compete with SPADE. The channel gate $M_c$ follows CBAM \cite{cbam}; the spatial gate $M_s$ conditions on the upsampled $\mathbf{p}$, directing capacity toward scattering-critical regions such as building edges and point targets. SAA is inserted once at the bottleneck, before the SPADE decoder, and adds only ${\sim}0.1\%$ of generator parameters.

\subsection{Training Objectives}

The generator is trained with a least-squares adversarial loss $\mathcal{L}_{\text{adv}}$, $\ell_1$ identity loss $\mathcal{L}_{\text{idt}}$, and $\ell_1$ cycle-consistency loss $\mathcal{L}_{\text{cyc}}$ for unpaired bidirectional translation. We additionally enforce grayscale consistency to prevent inter-channel drift in the SAR output:
\begin{equation}
    \mathcal{L}_{\text{gray}} = \|\hat{I}_{R} - \hat{I}_{G}\|_1 + \|\hat{I}_{G} - \hat{I}_{B}\|_1.
    \label{eq:gray}
\end{equation}
This term is applied identically to all methods, isolating the effect of the scattering-aware modules. The total objective is:
\begin{equation}
    \mathcal{L}_G = \lambda_{\text{adv}}\mathcal{L}_{\text{adv}} + \lambda_{\text{cyc}}\mathcal{L}_{\text{cyc}} + \lambda_{\text{idt}}\mathcal{L}_{\text{idt}} + \lambda_{\text{gray}}\mathcal{L}_{\text{gray}}
    \label{eq:total}
\end{equation}
with $\lambda_{\text{adv}}=1$, $\lambda_{\text{cyc}}=10$, $\lambda_{\text{idt}}=10$, $\lambda_{\text{gray}}=1$. The scattering prior enters through the architecture (SPADE/SAA conditioning on $\mathbf{p}$), not explicit physics losses; speckle-, frequency-, and scattering-classification losses were explored but proved ineffective (see also Section~\ref{sec:speckle_discuss}).

\section{Experiments}

\begin{figure*}[!t]
\centerline{\includegraphics[width=0.82\textwidth]{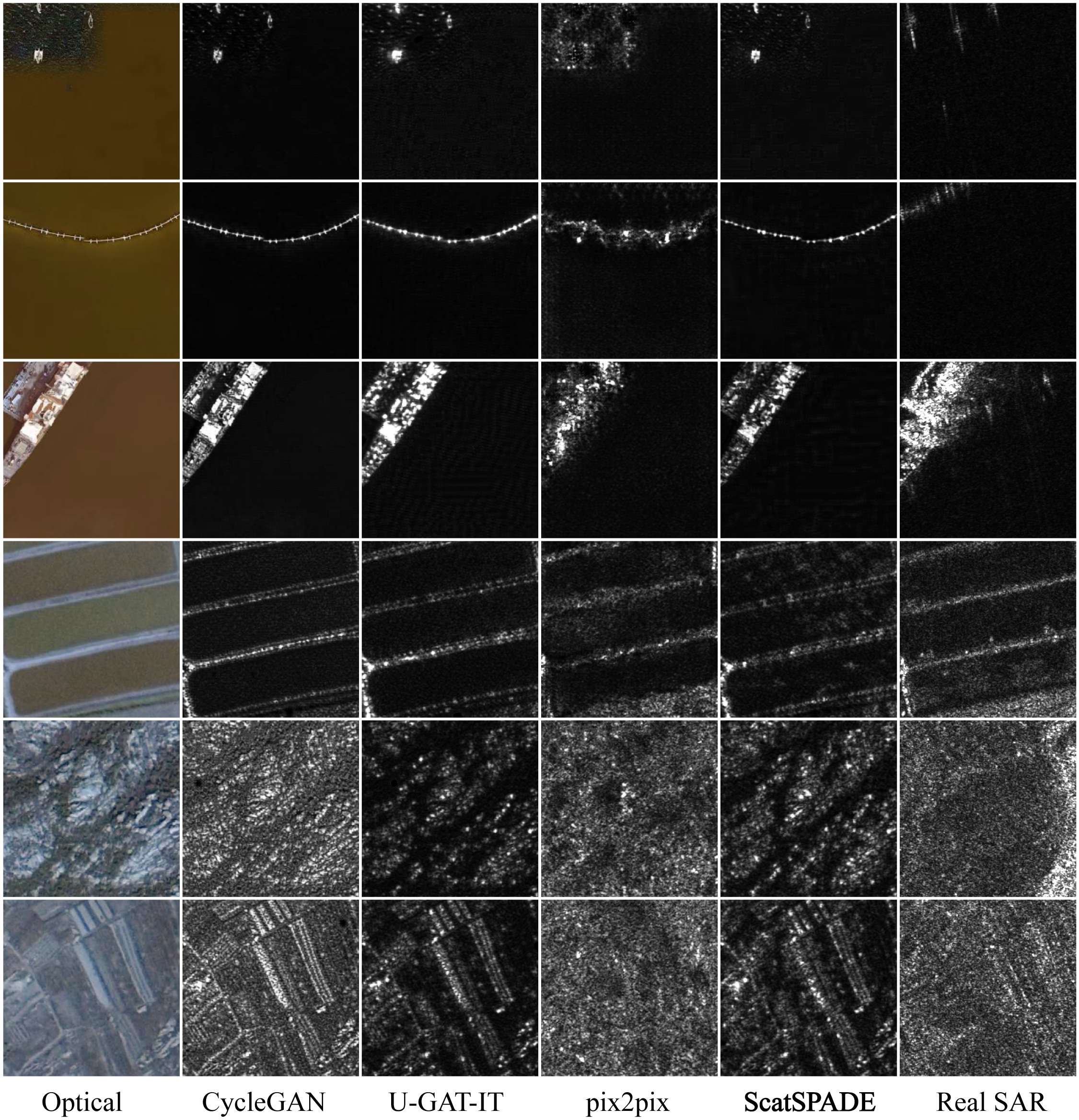}}
\caption{Qualitative comparison of generated SAR images. From left to right: input optical, CycleGAN, U-GAT-IT, pix2pix (paired upper bound), ScatSPADE (ours), and real SAR references. All generated columns are shown after grayscale projection for fair visual comparison.}
\label{fig:qualitative}
\end{figure*}

\subsection{Experimental Setup}

\textit{Dataset.} We evaluate on the QXS-SAROPT dataset \cite{qxs}, which contains 20{,}000 co-registered optical-SAR pairs at $256\times256$ resolution (GaoFen-3 C-band SAR and Google Earth optical), covering port cities with diverse scatterer types. We split the data into 16{,}000 training and 4{,}000 testing pairs. During training, the pairing is deliberately ignored (independent shuffled loaders), preserving the unpaired regime. During testing, the pairing is exploited to enable pixel-level evaluation.

\textit{Implementation.} The generator uses 4 residual blocks with a base channel of 64. The discriminator is a $70\times70$ PatchGAN \cite{ref27} with spectral normalization \cite{ref39}. All methods are trained for $10^5$ iterations using Adam ($\beta{=}(0.5,0.999)$, lr $10^{-4}$, batch size 2) on a single RTX 4090. CycleGAN and pix2pix use their official public implementations under identical training budgets; no method-specific hyperparameter tuning is applied. SPADE and SAA together add approximately $25\%$ more parameters than the U-GAT-IT generator, with SAA alone contributing less than $0.1\%$. All methods share an identical random seed and data split, giving a controlled comparison.

\textit{Metrics.} We adopt three categories of metrics. For \emph{distributional quality}, FID \cite{ref44} and KID \cite{ref45} measure the statistical distance between generated and real SAR in Inception-v3 feature space. For \emph{radiometric fidelity}, $|\Delta\mu|$ measures global-intensity deviation and $|\Delta\text{ENL}|=|\text{ENL}_{\text{gen}}-\text{ENL}_{\text{real}}|$ measures deviation of the equivalent number of looks, $\text{ENL}=\mu^2/\sigma^2$ computed on the three $64\times64$ sliding-window patches of lowest internal variance (the most homogeneous regions) per image (real-SAR ENL $\approx 3.33$); and $|\Delta\text{ISLR}|$ (integrated side-lobe-ratio deviation, a SAR-specific point-target sharpness metric). For \emph{structural fidelity}, PSNR, SSIM, and LPIPS are computed on 4{,}000 paired test images. We note that SAR speckle is a stochastic realization: the same scene admits many valid SAR appearances, so pixel-level PSNR/SSIM are intrinsically suppressed regardless of method quality. We additionally report $|R\text{-}B|$ (mean absolute red-blue channel difference) to verify grayscale correctness.

\subsection{Comparison with Baseline Methods}

We compare against three unpaired baselines (CycleGAN \cite{ref16}, CUT \cite{ref47}, U-GAT-IT \cite{ref31}) and include pix2pix \cite{ref27} as a paired upper bound. Table~\ref{tab:comparison} reports all metrics.

\begin{table*}[!t]
\caption{Quantitative comparison on QXS-SAROPT (single seed, identical for all methods). Best \emph{unpaired} result in \textbf{bold}. FID/KID ($\downarrow$); $|\Delta\mu|$, $|\Delta\text{ENL}|$, $|\Delta\text{ISLR}|$ ($\downarrow$); PSNR/SSIM ($\uparrow$); LPIPS ($\downarrow$); $|R\text{-}B|$ ($\downarrow$).}
\label{tab:comparison}
\centering
\footnotesize
\setlength{\tabcolsep}{4.5pt}
\begin{tabular}{lccccccccc}
\toprule
& \multicolumn{2}{c}{Distributional $\downarrow$} & \multicolumn{3}{c}{Radiometric $\downarrow$} & \multicolumn{3}{c}{Structural} & Color $\downarrow$ \\
\cmidrule(lr){2-3}\cmidrule(lr){4-6}\cmidrule(lr){7-9}\cmidrule(lr){10-10}
\textbf{Method} & FID & KID & $|\Delta\mu|$ & $|\Delta\text{ENL}|$ & $|\Delta\text{ISLR}|$ & PSNR$\uparrow$ & SSIM$\uparrow$ & LPIPS$\downarrow$ & $|R\text{-}B|$ \\
\midrule
\multicolumn{10}{l}{\textit{Unpaired}} \\
CycleGAN \cite{ref16}     & 129.96 & 0.098 & 0.046 & 8.45 & 0.18 & 9.93 & 0.044 & 0.495 & 2.45 \\
CUT \cite{ref47}          & 153.05 & 0.134 & 0.139 & 11.10 & 0.95 & \textbf{11.67} & \textbf{0.093} & 0.570 & \textbf{0.16} \\
U-GAT-IT \cite{ref31}     & 52.21 & 0.033 & 0.052 & \textbf{0.87} & \textbf{0.02} & 10.19 & 0.059 & \textbf{0.489} & 0.76 \\
\textbf{ScatSPADE (ours)} & \textbf{22.62} & \textbf{0.010} & \textbf{0.013} & 1.12 & 0.09 & 9.99 & 0.052 & \textbf{0.489} & 0.94 \\
\midrule
\multicolumn{10}{l}{\textit{Paired upper bound}} \\
pix2pix \cite{ref27}      & 62.92 & 0.052 & 0.041 & 0.27 & 0.39 & 12.67 & 0.099 & 0.388 & 1.89 \\
\bottomrule
\end{tabular}
\end{table*}

ScatSPADE achieves the lowest FID ($22.62$) and KID ($0.010$) among all methods, including the paired pix2pix upper bound ($62.92$), a $57\%$ reduction over the U-GAT-IT base architecture ($52.21$) and $64\%$ over pix2pix. Scattering-aware modulation thus produces SAR distributions closer to real data than every baseline, including one with full paired supervision, because conditioning $\gamma$ and $\beta$ on the scattering map yields region-specific intensity scaling that matches the bimodal brightness of real SAR.

A revealing pattern emerges across metrics. U-GAT-IT attains marginally higher PSNR ($10.19$ vs.\ $9.99$) yet far higher FID ($52.21$ vs.\ $22.62$); CUT is the extreme case, with the highest PSNR/SSIM of any unpaired method yet the worst realism (FID $153.05$), because contrastive learning preserves source structure but fails to adopt SAR statistics. Pixel-level metrics thus reward copying the input layout rather than synthesizing the stochastic speckle that defines SAR, and are intrinsically suppressed by speckle stochasticity. Distributional and perceptual metrics are therefore the more reliable realism criteria; ScatSPADE targets this realism and accepts lower pixel-level fidelity as a deliberate tradeoff.

On perceptual quality, ScatSPADE attains LPIPS $0.489$ (on par with U-GAT-IT), and its $|R\text{-}B|$ of $0.94$ indicates properly grayscale output, well below CycleGAN's color cast of $2.45$.

Fig.~\ref{fig:qualitative} provides visual confirmation: ScatSPADE generates speckle textures and point-like returns resembling real SAR, whereas CycleGAN and U-GAT-IT produce overly smooth outputs lacking SAR-specific texture.

\subsection{Ablation Study}

\begin{table}[t]
\caption{Ablation of the two scattering-conditioned modules (single seed). Baseline = U-GAT-IT. ``Full'' is the complete ScatSPADE model in Table~\ref{tab:comparison}. ``optical cond.'' replaces the scattering map with downsampled optical channels, a non-scattering spatial prior that isolates scattering-specificity. The standalone SAA row uses the gate form, whereas the full model uses the residual aggregation of \eqref{eq:saa}. FID/LPIPS/$|\Delta\text{ENL}|$ ($\downarrow$); PSNR ($\uparrow$). Best in \textbf{bold}.}
\label{tab:ablation}
\centering
\footnotesize
\setlength{\tabcolsep}{5pt}
\begin{tabular}{lcccc}
\toprule
\textbf{Configuration} & \textbf{FID}$\downarrow$ & $|\Delta\text{ENL}|\downarrow$ & \textbf{PSNR}$\uparrow$ & \textbf{LPIPS}$\downarrow$ \\
\midrule
Baseline (U-GAT-IT)              & 52.21 & 0.87 & 10.19 & 0.489 \\
+ SPADE (scatter cond.)         & 23.67 & 1.30 & 9.81 & 0.490 \\
+ SPADE (optical cond.)         & 29.55 & 0.60 & \textbf{10.70} & \textbf{0.465} \\
+ SAA only (gate)               & 44.07 & \textbf{0.58} & 10.38 & 0.481 \\
\textbf{Full (ours)}                  & \textbf{22.62} & 1.12 & 9.99 & 0.489 \\
\bottomrule
\end{tabular}
\end{table}

Table~\ref{tab:ablation} isolates each module's contribution. Adding scatter-conditioned SPADE cuts FID from $52.21$ to $23.67$, a $55\%$ reduction, confirming that scattering-conditioned affine parameters let the decoder produce region-specific intensity profiles: strong-scatterer regions retain sharp peaks while distributed-scatterer regions receive smoother modulation matching real SAR's granular speckle.

\textbf{Conditioning-source control.} Replacing the scattering map with downsampled optical channels---a non-scattering spatial prior---raises FID to $29.55$, $20\%$ worse than scatter-conditioned SPADE ($23.67$). Most of the SPADE gain is thus spatially-adaptive normalization per se, but a meaningful $20\%$ is specifically attributable to the scattering map, which is not interchangeable with arbitrary spatial conditioning. (The optical control scores higher pixel-fidelity by over-smoothing less---the realism--fidelity tradeoff; we target realism.)

SAA alone improves FID to $44.07$ and attains the best radiometric match ($|\Delta\text{ENL}|=0.58$) and structural fidelity (PSNR $10.38$, LPIPS $0.481$), showing that directing representational capacity toward scattering-critical regions enhances local texture and perceptual quality.

Combining both modules yields the full model with the lowest FID of $22.62$---lower than either alone (SPADE $23.67$, SAA $44.07$). This is exactly what the residual aggregation in \eqref{eq:saa} targets: blending the gated feature with the identity through a learnable $s$ lets SAA refine bottleneck features without over-suppressing them, complementing rather than competing with SPADE. SPADE thus drives distributional realism while SAA enhances perceptual fidelity, together reaching the most realistic SAR distribution.

\begin{figure}[!t]
    \centering
    \includegraphics[width=\columnwidth]{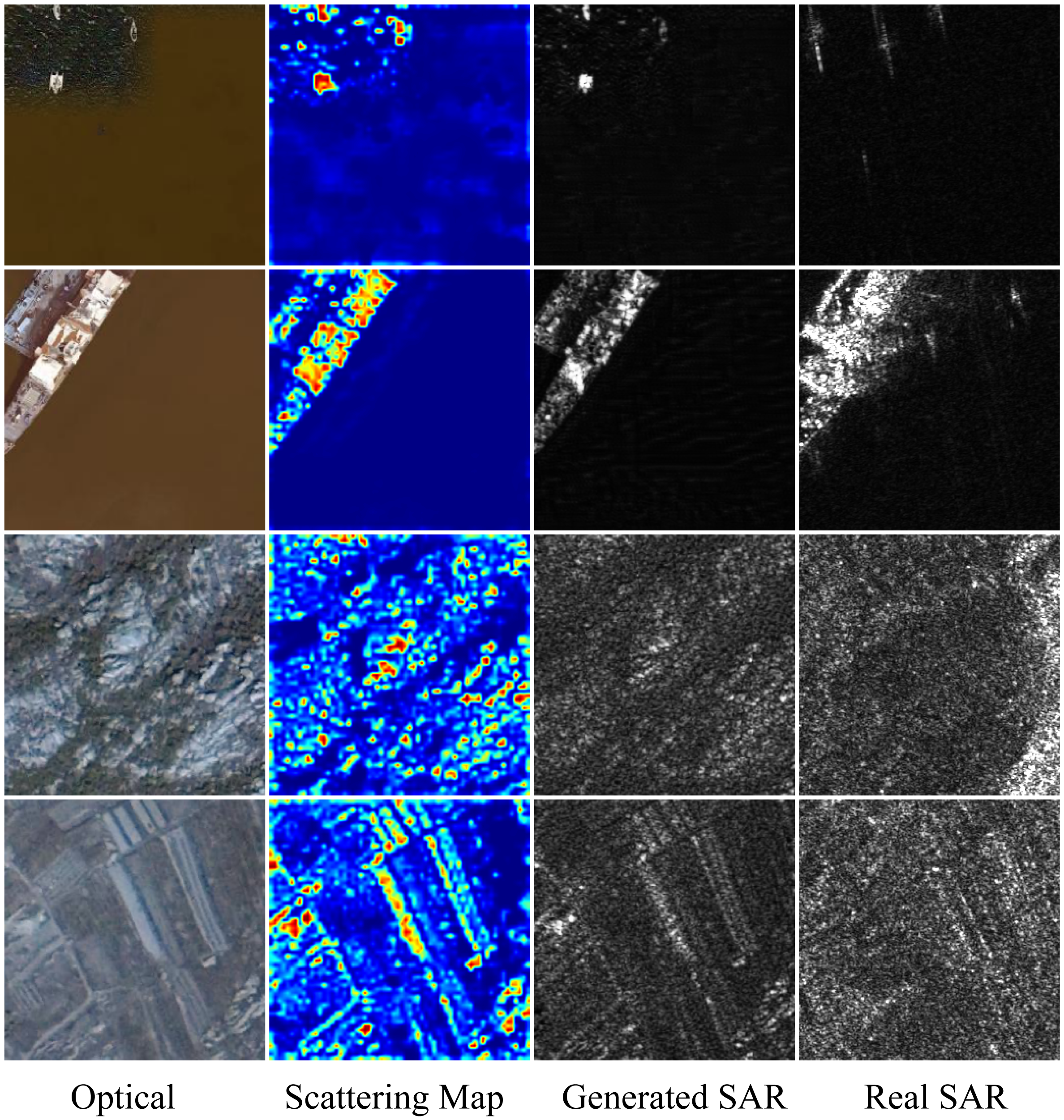}
    \caption{Scattering-conditioning visualization. Left-to-right per row: input optical, learned conditioning map (jet colormap), generated SAR, and real SAR reference. The map tracks scene structure---bright (red) responses concentrate on buildings and metallic structures that act as strong point scatterers, while dark (blue) regions mark distributed scatterers such as vegetation and water---and drives both SPADE and SAA.}
    \label{fig:scatter_map}
\end{figure}

Fig.~\ref{fig:scatter_map} visualizes the learned maps, whose strong-/distributed-scatterer responses (detailed in the caption) confirm that the bottleneck prior tracks scene structure and drives the region-adaptive modulation in SPADE and SAA.

\subsection{Discussion: Speckle Realism and Output-Stage Injection}
\label{sec:speckle_discuss}

Although ScatSPADE attains the most realistic SAR distribution, its speckle remains smoother than real SAR ($|\Delta\text{ENL}| = 1.12$, ENL $4.45$ vs.\ real $3.33$). To test whether injecting physically faithful speckle closes this gap, we modeled SAR intensity as $I = R \cdot S$, $S \sim \mathrm{Gamma}(L,L)$ at $L{=}3.33$, and trained identically otherwise; this collapsed catastrophically (FID $22.6\to239.3$, $|\Delta\mu|$ $0.013\to0.262$, output mean $\approx 0.01$ vs.\ $0.33$ real, output ENL $0.64$).

We attribute this to three coupled causes: real SAR speckle is scene-modulated rather than i.i.d.\ Gamma noise, so the discriminator rejects the uncontrollable texture and drives the generator to suppress $R$ toward $R\cdot S\to 0$; the output gradient scales with the high-variance speckle $S$ and vanishes near zero; and identity/cycle losses acquire an irreducible stochastic residual. SAR realism is thus governed by multiplicative speckle, yet output-stage noise injection is incompatible with adversarial training. Our structural alternative learns scattering-conditioned statistics through SPADE and SAA rather than adding output noise, so the residual speckle smoothness is an inherent cost of the structurally learnable route, likely requiring latent-space speckle modeling in future work.

\section{Conclusion}

In this paper, we proposed ScatSPADE, a scattering-aware framework for unpaired optical-to-SAR translation that injects a self-supervised scattering prior into the generator through two complementary modules---Scattering-Aware SPADE, which replaces uniform normalization with spatially adaptive feature statistics, and a Scattering-Aware Attention block, which re-weights bottleneck features toward scattering-critical regions---both driven by a single bottleneck scattering map and requiring no manual annotation.

Experiments on the QXS-SAROPT benchmark show that this design reduces FID by $57\%$ over the U-GAT-IT base architecture (to $22.62$) and by $64\%$ over the paired pix2pix upper bound, while attaining the lowest KID and among the lowest LPIPS of the unpaired methods. Ablation analysis confirms that each module contributes independently and that their combination is complementary rather than redundant.

Several limitations remain. The synthesized speckle texture is smoother than that of real SAR; as discussed in Section~\ref{sec:speckle_discuss}, we verified that direct output-stage speckle injection collapses the generator, indicating that this gap is not closable by naive injection and instead calls for scene-conditioned or latent-space speckle modeling. The binary scattering map is a coarse, self-supervised spatial conditioner rather than a calibrated scattering classifier; we adopt it for its realism benefit, but physically faithful scattering supervision is left to future work. Future work will also explore richer scattering decompositions, latent-space speckle modeling, multi-dataset generalization, and downstream validation through SAR target detection with augmented training data.

\bibliographystyle{IEEEtran}
\bibliography{references}

\end{document}